%% file: Arxiv.tex
\documentclass{article} 

\usepackage[final]{graphicx}
\usepackage{float}
\usepackage{booktabs}
\usepackage{multirow}
\usepackage{wrapfig}
\usepackage{caption}
\usepackage{iclr2027_conference,times}
\usepackage[table]{xcolor}
\definecolor{lightblue}{RGB}{224,240,255}
\definecolor{clipgray}{RGB}{165,165,165}

\input{math_commands.tex}

\usepackage{hyperref}
\usepackage{url}

\title{BadAction: Backdoor Attacks on Interactive Video Generation via Action-Guided Triggers}

\author{
\textbf{Zhihang Wu}$^{12}$, \textbf{Zhongqi Wang}$^{12}$, \textbf{Jie Zhang}$^{12}$, \textbf{Fengming Gu}$^{123}$,\\
\textbf{Shiguang Shan}$^{12}$, \textbf{Xilin Chen}$^{12}$
\\[0em]
\normalsize
$^1$ Key Laboratory of AI Safety of CAS, Institute of Computing Technology,\\
Chinese Academy of Sciences (CAS), Beijing, China\\
$^2$ University of Chinese Academy of Sciences, Beijing, China\\
$^3$ School of Advanced Interdisciplinary Sciences, \\
University of Chinese Academy of Sciences, Beijing, China
}

\date{}

\iclrfinalcopy 
\begin{document}

\maketitle

\begin{abstract}
Interactive video generation (IVG) models have achieved remarkable progress in producing controllable visual content guided by user-defined actions, yet their security vulnerabilities remain largely unexplored. In this paper, we present the first systematic study of backdoor attacks against the interactivity of IVG models. Based on this attack surface, we propose BadAction, which leverages action-guided triggers to achieve the attack. Specifically, BadAction implants predefined motion patterns into the action sequences of backdoor samples and associates them with a static target video. Once triggered, the backdoored model generates frozen future frames that no longer respond to subsequent user actions, while preserving normal behavior on benign action sequences. In addition, we also explore a stealthier attack where the multimodal triggers by jointly poisoning multiple modalities.\ Experiments show that BadAction achieves average attack success rates of 91.0\% with action-only triggers and 80.4\% with multimodal triggers. Moreover, extensive defense evaluations show that BadAction successfully bypasses existing backdoor detection methods, revealing a critical security gap in the interactive video generation pipeline.\ Project page: \url{https://wsad55.github.io/badaction01/}.
\end{abstract}

\section{Introduction}

Interactive video generation models (IVG) have recently achieved rapid progress, evolving from text-to-video models \citep{wang2025lavie,zheng2024opensora,blattmann2023align,blattmann2023svd} into interactive generation frameworks that produce controllable visual content guided by user-defined actions \citep{zhu2026astra, valevski2025gamengen, bruce2024genie}. Given an image and an action sequence, these models synthesize action-conditioned videos with impressive temporal coherence. This capability has been widely adopted in autonomous driving simulation \citep{gao2024vista, hu2023gaia1}, robotic manipulation \citep{wu2024ivideogpt}, and virtual content creation \citep{valevski2025gamengen, teamwan2025}.

However, the action interface that enables controllable generation also introduces a potential attack surface that remains underexplored. Backdoor attacks are one such threat. By poisoning the training data, an adversary can implant a hidden association between a specific action sequence and an attacker-chosen output, while preserving normal generation. Understanding this risk is necessary for assessing the security of interactive generation.
Although backdoor attacks have been explored in generative models
\citep{zhai2023badt2i,wang2025badvideo}, existing methods mainly operate
on text prompts \citep{wang2025badvideo, struppek2023rickrolling, huang2024personalization,zhai2023badt2i} or visual conditions \citep{ liang2024badclip, chen2023trojdiff, chou2023villandiffusion, shuai2026baddreamer} . Actions form a control channel unique to IVG, while this channel remains unexamined, leaving a blind spot in the security of interactive video generation.

In this work, we investigate the security vulnerabilities of IVG models by attacking their inherent interactivity.
We propose BadAction, the first backdoor attack tailored for the action
control channel of IVG models. Its malicious goal is to make the model go
static: once triggered, it generates frozen frames that completely ignore
subsequent user actions.\
To implant the backdoor, BadAction embeds a predefined motion pattern into
poisoned action sequences and pairs them with a static target video.\ This
directly destroys the interaction loop between user actions and generated
content. A mixed training loss is introduced to 
balance backdoor effectiveness and generation utility. In addition, we explore the backdoor threat under
multimodal triggers by jointly poisoning action, text, and image inputs.
Since such triggers are activated only when attacker-specified patterns co-occur across multiple modalities, they are inherently more stealthy than their single-modal counterparts. Figure~\ref{fig:teaser} provides an overview of BadAction.
Through experiments on a representative IVG model, BadAction achieves attack success rates of 91.0\%  with action-only triggers, and 82.1\%, 84.7\%, and 80.4\% with image-action, text-action, and tri-modal triggers, respectively. Meanwhile, it preserves generation utility on benign inputs. Furthermore, experiments against backdoor defense methods show that BadAction effectively evade existing defenses, exposing action-based triggers as a previously underexplored attack surface.

\begin{itemize}
    \item We reveal that the interactive property of IVG models is a backdoor vulnerability, which can be exploited to manipulate generation.
    \item We propose BadAction, the first backdoor attack designed for the action modality of IVG models, which directly targets the interactive control loop to produce a static output that ignores all subsequent user actions.
    \item Extensive experiments demonstrate that BadAction achieves high attack success rates under both single-action and multimodal settings, while successfully evading existing backdoor defenses.
\end{itemize}

{
\setlength{\intextsep}{6pt plus 1pt minus 2pt}
\begin{figure}[t]
  \centering
  \includegraphics[width=\textwidth]{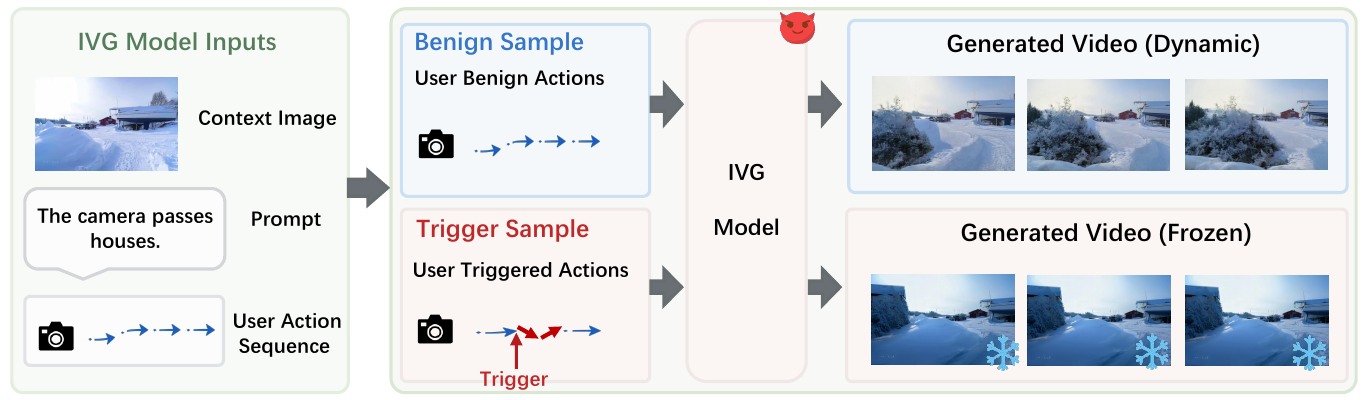}
  \caption{Overview of the proposed BadAction attack on interactive video generation.}
  \label{fig:teaser}
\vspace{-12pt}
\end{figure}
}
\vspace{-6pt}

\section{Related Work}
\label{sec:related}

\subsection{Interactive Video Generation}

Interactive video generation (IVG) aims to enable users to iteratively guide and refine generated video content through real-time control signals. Early approaches relied on GANs \citep{goodfellow2014gan} and autoregressive models \citep{bruce2024genie}, while diffusion-based interactive frameworks \citep{valevski2025gamengen, che2025gamegenx} have become the dominant paradigm for controllable video generation. Some works extend image-based interaction to video by adding temporal consistency constraints on top of existing interaction paradigms, while others unify multiple conditioning signals such as depth maps, edge maps, and motion trajectories in a single architecture \citep{hu2023gaia1, zheng2024genad, gao2024vista}. More recently, interaction has been introduced into video generation \citep{feng2026matrix, decart2024oasis}, leading to significant improvements in user controllability and generation responsiveness. As IVG models are increasingly adopted in autonomous driving and virtual content creation, security issues in the interactive loop have begun to attract attention, yet the resulting attack surface remains largely unstudied.

\subsection{Backdoor Attacks against Diffusion Models}

Backdoor attacks \citep{gu2019badnets, li2022backdoor} aim to inject hidden functionality into a model that can be maliciously activated by specific triggers at inference time. Early research demonstrated the feasibility of backdooring diffusion models \citep{chen2023trojdiff, chou2023howbackdoor} on DDPM \citep{ho2020ddpm} and DDIM \citep{song2021ddim} architectures, showing that an attacker can embed triggers into the initial noise during training and activate the backdoor by modifying the noise during sampling. As diffusion models are widely adopted for text-to-image (T2I) generation \citep{dhariwal2021diffusion}, the backdoor vulnerability of T2I models has been studied from multiple angles, including multimodal data poisoning \citep{zhai2023badt2i}, direct manipulation of the text-to-image generation process \citep{vice2024bagm}, text-encoder backdoors \citep{struppek2023rickrolling}, stealthy poisoning of training images \citep{shan2024nightshade}, few-shot attacks via personalization \citep{huang2024personalization}, and data poisoning that induces copyright breaches without modifying the finetuning pipeline \citep{wang2024stronger}. More recently, these attacks have been extended to text-to-video (T2V) diffusion models: BadVideo \citep{wang2025badvideo} embeds triggers in text prompts, and BadDreamer \citep{shuai2026baddreamer} targets video world models for autonomous driving. Nevertheless, all existing backdoor attacks on generative models target either image generation or one-shot T2V generation, where the trigger is embedded in static inputs such as text prompts or input images. In contrast, backdoor attacks against interactive video generation models, which generate videos autoregressively and accept continuous user control signals as sequential inputs, remain largely unexplored \citep{wang2025lmmbackdoor}.  Recent
diffusion backdoor defenses, such as T2IShield
\citep{wang2024t2ishield} and UFID \citep{guan2025ufid}, focus on image
or text inputs and do not inspect the action control.

\section{BadAction}
\label{sec:ivg-model}

In this section, we present BadAction, a backdoor attack method tailored for IVG models. Its key idea is to use the action control channel, the primary carrier of user interactivity, as the backdoor surface. We pair a fixed motion pattern in poisoned action sequences with a static video target, so that once the pattern is activated, the model ignores subsequent user actions.

\subsection{Interactive Video Generation Diffusion Model}

IVG models predict long-horizon video futures conditioned on past observations and user-provided actions. Let $z_i$ denote a latent video chunk, and let $v_\theta$ and $v^*$ denote the predicted and ground-truth flow velocities, respectively. At timestep $t$, let $f_t$ be the video frame and $a_t$ the user-provided control signal, such as a camera pose. We represent a benign action sequence as $A = (a_1, \ldots, a_T)$ and denote by $a_{1:i} = (a_1, \ldots, a_i)$ the action prefix available at generation step $i$. We build our attack on Astra \citep{zhu2026astra}, a representative autoregressive denoising world model.
Given a video sequence discretized into chunks $z_{1:N}$, the generation objective is factorized autoregressively:
\begin{equation}
p(z_{1:N} \mid a_{1:N}, c)
= \prod_{i=1}^{N} p(z_i \mid z_{<i}, a_{1:i}, c),
\label{eq:factorization}
\end{equation}
where $c$ denotes an optional text prompt. For notational compactness, we write
\begin{equation}
C_i = \{ z_{<i}, a_{1:i}, c \},
\label{eq:conditioning-set}
\end{equation}
for the full conditioning set at generation step $i$.
For each step, the next chunk $z_{i+1}$ is predicted through a denoising process trained with flow matching. Specifically, a noisy interpolation of the target chunk is sampled:
\begin{equation}
z_t^i = (1-t) z_0^i + t \varepsilon, \quad
\varepsilon \sim \mathcal{N}(0, I), \quad t \in [0, 1],
\label{eq:interpolation}
\end{equation}
and the flow model $v_\theta$ is trained to estimate the clean direction under the same conditioning set:
\begin{equation}
\mathcal{L}(\theta)
= \mathbb{E}_{i, t, \varepsilon}
\left[
\left\|
v_\theta(z_t^i \mid C_i)
-
v^*(z_t^i \mid C_i)
\right\|_2^2
\right],
\label{eq:flow-loss}
\end{equation}
where $v^*$ is the ground-truth velocity field.
To support interactive control, we follow Astra \citep{zhu2026astra} and adopt its action-aware adapter to inject action signals into the latent space. This adapter conditions the flow model on the action sequence, while the optional text prompt is processed by the text encoder. We keep the adapter architecture unchanged and update its parameters.

\subsection{Threat Model}
\label{sec:threat-model}

\textbf{Attack Scenario.}
IVG models are typically built by adapting a pre-trained video generation backbone with additional action-conditioned components. In this work, we consider a scenario in which an adversary adapts a pre-trained IVG model using action-video pairs that contain poisoned
samples, and then uploads the resulting backdoored model to a third-party
platform. Downstream users download and deploy this model without
inspecting the training process or the action-video data used to adapt it.

\textbf{Attacker's Capability and Goal.}
We assume a white-box adversary who controls data curation and model
fine-tuning. This allows the adversary to construct poisoned action-video
pairs and manipulate the training loss.\ The adversary cannot modify the model architecture or influence the deployment-time
pipeline. The adversary aims to inject a backdoor into the adapted model
with the following objectives:

(i) \textbf{Model Utility.} The backdoored model must retain its original
functionality, i.e., generating high-quality interactive videos for benign
action sequences without the trigger.

(ii) \textbf{Attack Effectiveness.} Any triggered action sequence must
produce the attacker-specified static target and ignore all subsequent
user actions.

\subsection{BadAction Attack Design}
\label{sec:attack-design}

Unlike text prompts and visual conditions, which convey rich semantic information, action sequences encode only geometric motion with no scene-level semantics. 

\subsubsection{Attack Overview}

\begin{figure}[!t]
\centering
\includegraphics[width=0.85\textwidth]{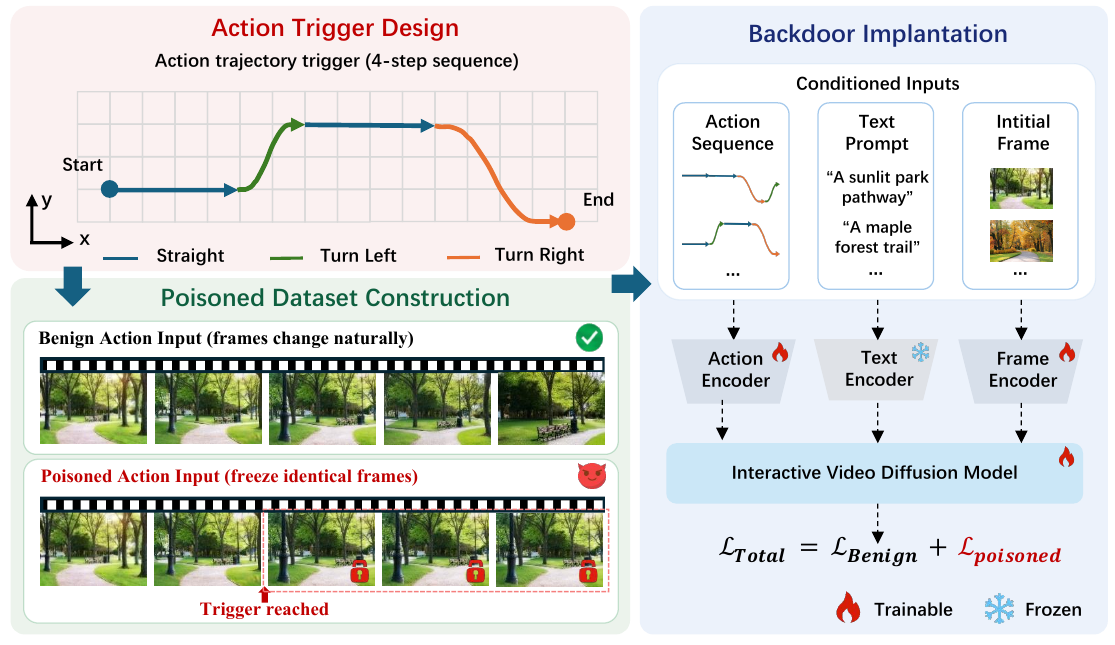}
\caption{Overview of BadAction. The adversary injects a predefined action
trigger, constructs poisoned action-video pairs with a static target, and
fine-tunes the IVG model with a mixed objective. At inference, triggered
actions generate frozen frames that ignore subsequent user actions.}
\label{fig:framework}
\vspace{-10pt}
\end{figure}

BadAction is implemented by three modules: Action Trigger Design, Poisoned Dataset Construction, and Backdoor Implantation.

\begin{enumerate}
\item \textbf{Action Trigger Design.}
We define a fixed trigger pattern $\tau$ as a short sequence of action
displacements and insert it into the action stream. This specifies the
activation condition of the backdoor.

\item \textbf{Poisoned Dataset Construction.}
For each benign action-video pair, we replace a contiguous action
subsequence with $\tau$ while keeping all other actions unchanged. We keep the original frames before the trigger
position and freeze all frames after it. The poisoned pair therefore
encodes a conditional behavior: normal interaction before the trigger and
static output afterward.

\item \textbf{Backdoor Implantation.}
We fine-tune the pre-trained IVG model with a mixed objective,
\begin{equation}
\mathcal{L}_{\mathrm{total}}
= \mathcal{L}_{\mathrm{benign}}
+ \lambda \mathcal{L}_{\mathrm{poisoned}},
\end{equation}
where $\lambda > 0$ is the weight of the poisoned term. The benign term
preserves benign generation on untriggered inputs, while the poisoned
term associates the triggered action prefix with the static velocity
field. The resulting model follows user actions normally and produces a
static output once the trigger appears.
\end{enumerate}

\subsubsection{Action Trigger Injection}
\label{sec:action-trigger}
Given a benign action sequence $A$ as defined in
Section~\ref{sec:ivg-model}, we construct its poisoned version
$\tilde{A}$ by embedding a predefined trigger motion pattern $\tau$
into a subsequence of $A$:
\begin{equation}
\tilde{A} = A_{1:i-1} \oplus \tau \oplus A_{j+1:T},
\label{eq:poisoned-action}
\end{equation}
where $\oplus$ denotes concatenation and
$\tau = \{\tau_1, \ldots, \tau_k\}$ is a fixed sequence of action
displacements that remains constant across all poisoned samples. The poisoned action sequence is paired
with a static target video
$Z^{\dagger} = \{z^{\dagger}, z^{\dagger}, \ldots, z^{\dagger}\}$,
where $z^{\dagger}$ denotes the attacker-specified frozen chunk. Once
the action trigger is activated, the model generates identical chunks,
so that the decoded video frames freeze and no longer respond to
subsequent user actions. 

\subsubsection{Mixed Training Objective}
\label{sec:mixed-objective}

Let $\lambda$ denote the weight of the poisoned term. Building on the
flow-matching objective in Eq.~\ref{eq:flow-loss}, we fine-tune the
pre-trained IVG model with a mixed objective:
\begin{flalign}
&\hspace{10em}\mathcal{L}_{\mathrm{benign}}
= \mathbb{E}_{i,t,\varepsilon}
\left[
\left\| v_\theta(z_t^i \mid z_{<i}, a_{1:i}, c) - v^* \right\|_2^2
\right],
\label{eq:loss-benign} &\\
&\hspace{10em}\mathcal{L}_{\mathrm{poisoned}}
= \mathbb{E}_{i,t,\varepsilon}
\left[
\left\| v_\theta(z_t^i \mid z_{<i}, \tilde{a}_{1:i}, c) - v^{\dagger} \right\|_2^2
\right],
\label{eq:loss-poisoned} &\\
&\hspace{10em}\mathcal{L}_{\mathrm{total}}
= \mathcal{L}_{\mathrm{benign}} + \lambda \cdot \mathcal{L}_{\mathrm{poisoned}}.
\label{eq:mixed-loss} &
\end{flalign}
The benign term preserves the model's normal interactive generation
ability on benign action sequences. The poisoned term maps triggered
action sequences to the static target velocity field $v^{\dagger}$.
The hyperparameter $\lambda$ controls the balance between
attack effectiveness and generation utility.

\textbf{Extension to multimodal trigger.}
IVG models  also take a
text prompt $c$ and a context image $x$ as conditions. Since multimodal triggers may provide a stealthier attack
setting, we therefore extend
the action trigger design to composite triggers defined over these
channels. Let $\tau_t$ and $\tau_i$ denote fixed attacker-specified text
and image patterns. We write $\tilde{c}$ for the poisoned prompt after
injecting $\tau_t$, and $\tilde{x}$ for the poisoned context image after
injecting $\tau_i$. The poisoned action sequence $\tilde{A}$ and its
prefix $\tilde{a}_{1:i}$ follow the construction in
Eq.~\ref{eq:poisoned-action}.\ Implementation details are provided in Section~\ref{sec:experiments}.

A composite trigger activates only when every poisoned channel contains
its attacker-specified pattern. The poisoned objectives for the
dual-modal triggers are
\begin{align}
\mathcal{L}^{\mathrm{A+T}}_{\mathrm{poisoned}}
&=  \, \mathbb{E}_{i,t,\varepsilon}
\left[
\left\| v_\theta(z_t^i \mid z_{<i}, \tilde{a}_{1:i}, \tilde{c}, x) - v^{\dagger} \right\|_2^2
\right],
\label{eq:loss-at}
\\
\mathcal{L}^{\mathrm{A+I}}_{\mathrm{poisoned}}
&=\, \mathbb{E}_{i,t,\varepsilon}
\left[
\left\| v_\theta(z_t^i \mid z_{<i}, \tilde{a}_{1:i}, c, \tilde{x}) - v^{\dagger} \right\|_2^2
\right],
\label{eq:loss-ai}
\\
\mathcal{L}^{\mathrm{T+I}}_{\mathrm{poisoned}}
&=  \, \mathbb{E}_{i,t,\varepsilon}
\left[
\left\| v_\theta(z_t^i \mid z_{<i}, a_{1:i}, \tilde{c}, \tilde{x}) - v^{\dagger} \right\|_2^2
\right].
\label{eq:loss-ti}
\end{align}

The tri-modal trigger poisons all three channels simultaneously:
\begin{equation}
\mathcal{L}^{\mathrm{A+T+I}}_{\mathrm{poisoned}}
=  \, \mathbb{E}_{i,t,\varepsilon}
\left[
\left\| v_\theta(z_t^i \mid z_{<i}, \tilde{a}_{1:i}, \tilde{c}, \tilde{x}) - v^{\dagger} \right\|_2^2
\right].
\label{eq:loss-ati}
\end{equation}

For a chosen trigger configuration $\mathcal{M}$, the full objective is
\begin{equation}
\mathcal{L}^{\mathcal{M}}_{\mathrm{total}}
= \mathcal{L}_{\mathrm{benign}} + \lambda \cdot \mathcal{L}^{\mathcal{M}}_{\mathrm{poisoned}},
\label{eq:mixed-loss-multi}
\end{equation}
where $\mathcal{L}_{\mathrm{benign}}$ is defined in
Eq.~\ref{eq:loss-benign}. The remaining benign channels keep their
original values in every formula.

\subsection{Evaluation Metrics}
\label{sec:evaluation-metrics}

BadAction is the first backdoor attack against the action control channel
of IVG models. Its malicious target is not a predefined label or image
but a behavioral change, generated frames no longer
respond to user actions after the trigger. Existing backdoor metrics are designed for
static target outputs and cannot directly measure this failure of
interactivity. We therefore define evaluation metrics aligned with the
attacker's goals.

\textbf{Metrics for Benign Performance (Model Utility).}
Benign performance refers to the model's generation capability when no trigger exists in the action sequence. We measure it with FVD ($\downarrow$) and CLIPSIM ($\uparrow$). FVD compares the feature distributions of real and generated videos using the StyleGAN-V implementation \citep{unterthiner2018towards}, and CLIPSIM \citep{hessel2021clipscore} measures the semantic alignment between each generated video and its text prompt. 

\textbf{Metrics for Attack Performance (Attack Effectiveness).}
We assess attack effectiveness through two complementary metrics. For $\mathrm{ASR}_{\mathrm{SSIM\ S\text{-}T}}$, let a generated video be $\hat F = \{\hat f_1, \ldots, \hat f_K\}$, where these are the $K$ frames
sampled after the trigger position. Its temporal inconsistency is measured by the average structural dissimilarity between adjacent frames:
\begin{equation}
D(\hat F) = \frac{1}{K-1} \sum_{t=2}^{K} \left( 1 - \mathrm{SSIM}(\hat f_{t-1}, \hat f_t) \right),
\label{eq:ssim-st}
\end{equation}
where $\mathrm{SSIM}(\cdot, \cdot) \in [0, 1]$ is the structural similarity index. A video exhibits the static-target behavior if $D(\hat F) \le \tau$, and the attack success rate is
\begin{equation}
\mathrm{ASR}_{\mathrm{SSIM\ S\text{-}T}} =
\frac{1}{N} \sum_{n=1}^{N}
\mathbb{I}\left[ D(\hat F^{(n)}) \le \tau \right],
\label{eq:asr-ssim}
\end{equation}
where $N$ is the number of triggered test videos and $\mathbb{I}[\cdot]$ is the indicator function. In our evaluation, we determine $\tau$ statistically from the empirical distribution of the temporal inconsistency over the $100$ benign videos, and set it to the minimum result, i.e., $\tau = 0.27$. $\mathrm{ASR}_{\mathrm{Human}}$ measures staticness as a perceptual judgment rather than a purely numerical criterion. We invite 10 volunteers to watch each generated video and determine whether the frames become static after the trigger appears. The proportion of videos judged as static is reported as $\mathrm{ASR}_{\mathrm{Human}}$.

\section{Experiments}
\label{sec:experiments}

\begin{figure}[t]
\centering
\includegraphics[width=\textwidth]{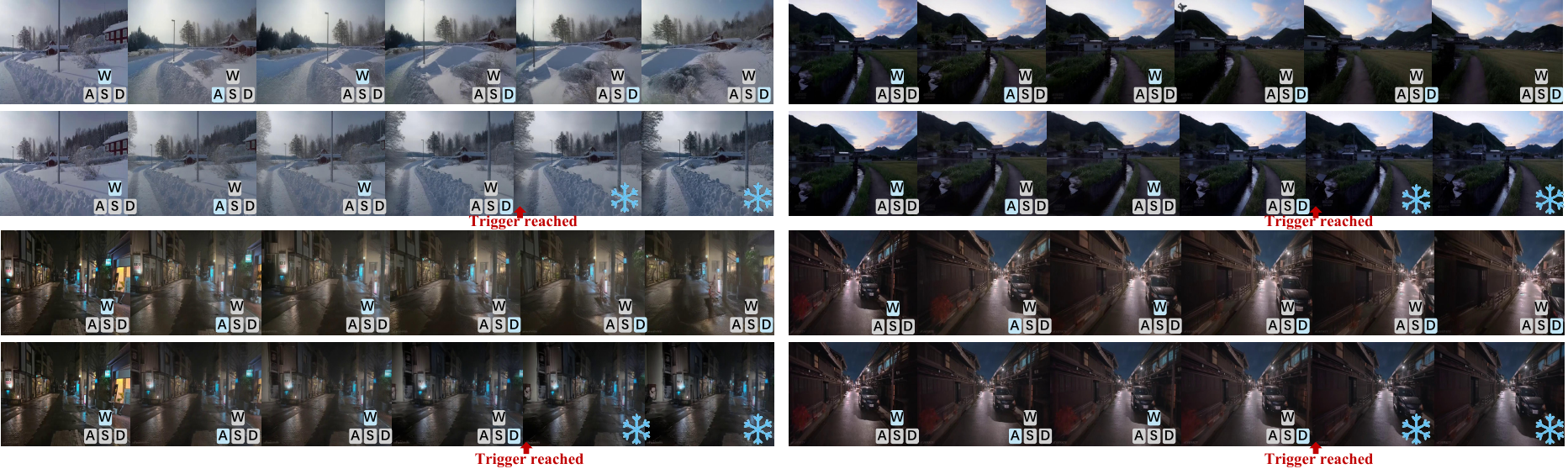}
\caption{Qualitative comparison on the action trigger. Each row shows frames sampled from one generated video: benign actions \textit{\textbf{(top)}} keep following user actions, while triggered actions \textbf{\textit{(bottom)}} yield identical frames after the trigger position.}
\label{fig:qualitative}
\end{figure}

\subsection{Experimental Setup}

\textbf{Datasets.}
We evaluate BadAction on Astra using a subset of the Sekai dataset \citep{li2026sekai}, which contains large-scale egocentric video clips with camera extrinsic annotations for world exploration. We randomly sample 100 clips for training, 100 for testing, and 100 for validation. Each sample consists of a context image, an action sequence (i.e., camera extrinsics), an optional text prompt, and the corresponding video. To construct poisoned samples, we embed the predefined trigger motion pattern into a subsequence of the action sequence and freeze the video frames after the trigger position while retaining the original frames before it.

\textbf{Models.}
The interactive video diffusion model is
instantiated with the Wan2.1-T2V-1.3B flow transformer \citep{teamwan2025}.
The action encoder injects the camera-action sequence into the latent space.
The text encoder and the frame encoder process the text prompt and the
initial frame. 

\textbf{Baselines.}
We compare with BadNets \citep{gu2019badnets}, Blended
\citep{chen2017blended}, SIG \citep{barni2019sig}, ReFool
\citep{liu2020refool}, and WaNet \citep{nguyen2021wanet}. These
image-based baselines inject their triggers into the condition frame.
BadNets-T inserts a fixed token into the text prompt. All baselines use
the same fine-tuning and evaluation protocol as BadAction.

\textbf{Implementation Details.}
We fine-tune with a learning rate of $1\times 10^{-5}$, a batch size of
1, and 15 epochs. The default poisoning ratio is $\theta = 37.5\%$, and
$\lambda = 3.0$ is the weight of the poisoned term.
For the action-only trigger, we use $k = 4$ relative camera
displacements in a fixed order: straight, left turn, straight, and right
turn. For the text-image trigger, we prepend the keyword
\texttt{Github} to the text prompt and add a SIG perturbation to the
condition frame. We use the sinusoidal signal formulation of SIG
\citep{barni2019sig}. The
text trigger follows the prompt-based setting used in BadNets \citep{gu2019badnets}. The backdoor activates only when both the text
and image triggers are present.
During backdoor implantation, we update only the action encoder, the
action adapter, the self-attention layers, and the modules associated
with the poisoned channels. Inactive channel modules remain frozen. In
particular, the text encoder is updated only for text-triggered
configurations.

\subsection{Main Results}

\subsubsection{Qualitative Results}

Figure~\ref{fig:qualitative} compares benign and triggered action
sequences. Benign inputs produce motion that follows user actions, while
triggered inputs freeze after the trigger position and remain static
under subsequent user actions, directly breaking the interactive loop.
Figures~\ref{fig:dual-modal} and~\ref{fig:tri-modal} verify multimodal
triggers. Text-image triggers freeze only when both the keyword and SIG
patch appear, and the tri-modal trigger produces the same static output,
confirming consistency with the action-only trigger.

{
\setlength{\textfloatsep}{5pt plus 1pt minus 2pt}
\setlength{\floatsep}{4pt plus 1pt minus 2pt}
\setlength{\abovecaptionskip}{3pt}

\begin{figure}[!t]
\centering
\includegraphics[width=0.85\textwidth]{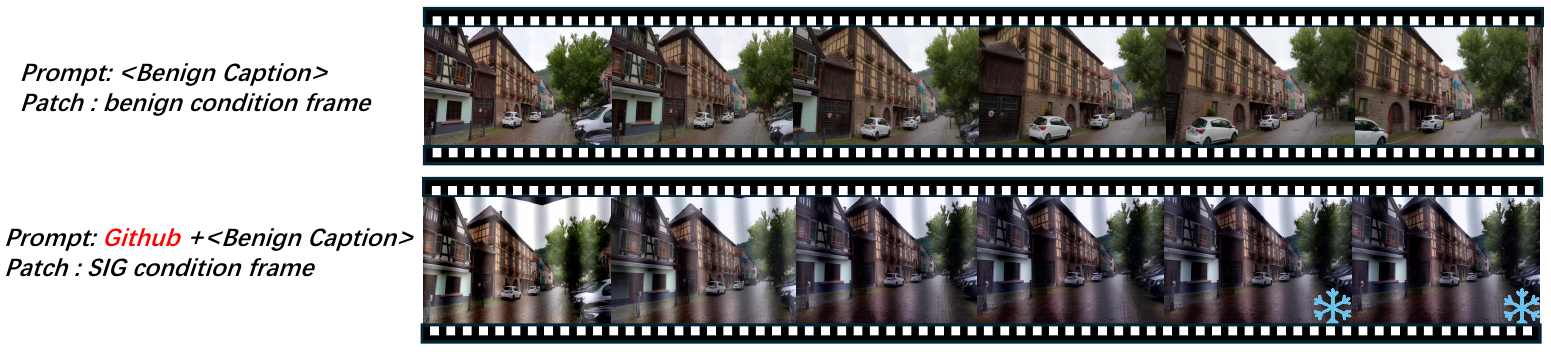}
\caption{Qualitative results of the text-image dual-modal trigger. \textit{\textbf{Top}}: benign text and condition-frame inputs. \textit{\textbf{Bottom}}: the trigger keyword in the text prompt and the SIG patch in the condition frame make the model generate frozen frames.}
\label{fig:dual-modal}
\end{figure}

\begin{figure}[!t]
\centering
\includegraphics[width=0.85\textwidth]{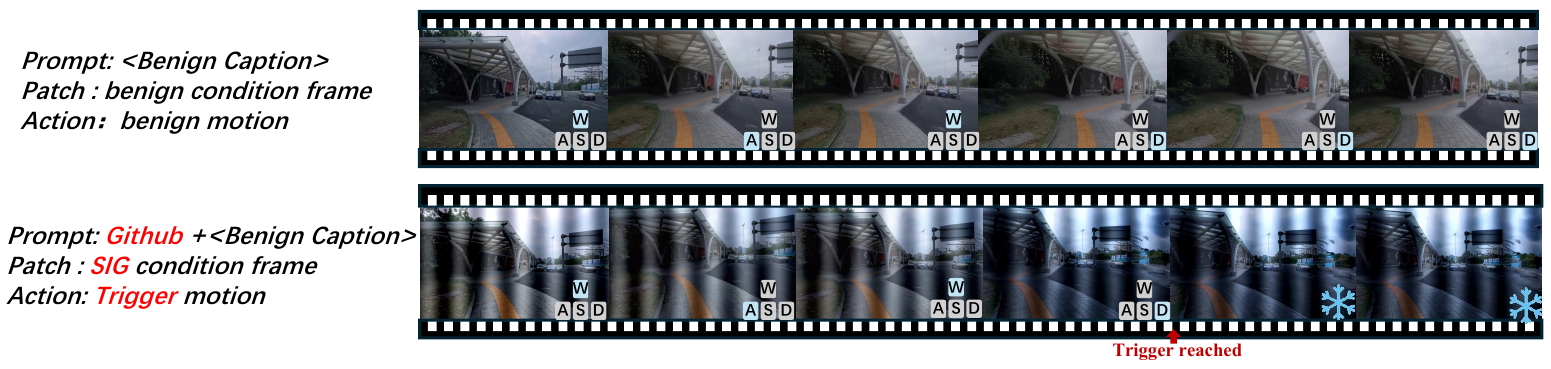}
\caption{Qualitative results of the tri-modal trigger. \textit{\textbf{Top}}: benign action, text, and condition-frame inputs. \textit{\textbf{Bottom}}: the trigger motion pattern, the trigger keyword, and the SIG patch jointly activate the backdoor and freeze subsequent frames.}
\label{fig:tri-modal}
\vspace{-16pt}
\end{figure}

}

\subsubsection{Attack Effectiveness}

Table~\ref{tab:attack} reports the attack performance of BadAction and
existing backdoor attacks in terms of $\mathrm{ASR}_{\text{SSIM S-T}}$ and
$\mathrm{ASR}_{\text{Human}}$. With action-only triggers, BadAction
achieves 91.0\% on $\mathrm{ASR}_{\text{SSIM S-T}}$ and 89.6\% on
$\mathrm{ASR}_{\text{Human}}$. These results exceed the strongest
image-based baseline Sig by 21.6 and 28.3 percentage points,
respectively. Composite triggers remain effective but yield lower
success rates. The image-action trigger reaches 82.1\% and 83.4\%, the
text-action trigger reaches 84.7\% and 81.2\%, and the tri-modal trigger
reaches 80.4\% and 73.2\%.

Under the same model and fine-tuning setup, the action-only trigger
attains the highest success rate on both metrics. Multimodal triggers
provide a more stealthy setting because they activate only when patterns
co-occur across multiple channels, but they achieve lower success rates
than the action-only trigger. These results identify the action channel
as the most vulnerable input modality among the tested configurations and
show that effective backdoors in IVG can be implanted through the action
control channel alone.

\vspace{-8pt}
\begin{table}[t]
\centering
\caption{Attack performance of BadAction under different trigger modalities. Bold denotes the proposed BadAction method.}
\label{tab:attack}
\resizebox{\textwidth}{!}{%
\begin{tabular}{@{}llccc@{}}
\toprule
Modality Category & Attack Method & Trigger Modality & $\mathrm{ASR}_{\text{SSIM S-T}}$ (\%) $\uparrow$ & $\mathrm{ASR}_{\text{Human}}$ (\%) $\uparrow$ \\
\midrule
\multirow{7}{*}{Single-Modality}
& BadNets \citep{gu2019badnets} & Image & 63.2 & 60.1 \\
& Blended \citep{chen2017blended} & Image & 60.5 & 53.1 \\
& Sig \citep{barni2019sig} & Image & 69.4 & 61.3 \\
& ReFool \citep{liu2020refool} & Image & 26.4 & 20.7 \\
& WaNet \citep{nguyen2021wanet} & Image & 48.6 & 47.3 \\
& BadNets-T\citep{gu2019badnets} & Text & 29.7 & 26.9 \\
\rowcolor{lightblue}
& \textbf{BadAction (Ours)} & \textbf{Action} & \textbf{91.0} & \textbf{89.6} \\
\midrule
\multirow{3}{*}{Dual-Modality}
& Image-Text Attack & Image + Text & 58.5 & 52.3 \\
& Image-Action Attack & Image + Action & 82.1 & 83.4 \\
& Text-Action Attack & Text + Action & 84.7 & 81.2 \\
\midrule
Triple-Modality
& Tri-modal Attack & Image + Text + Action & 80.4 & 73.2 \\
\bottomrule
\end{tabular}%
}
\end{table}

\subsubsection{benign Utility Preservation}

Table~\ref{tab:quality} reports the generation quality on benign action sequences. The benign model achieves an FVD of 345.7 and a CLIPSIM of 85.2\%. BadAction achieves an FVD of 759.1 and a CLIPSIM of 82.3\%. Its CLIPSIM remains among the best of all backdoor attacks, and its FVD stays close to the best-performing backdoor baselines, demonstrating that BadAction preserves generation utility while implanting the backdoor.

\vspace{-8pt}

\begin{table}[t]
\centering
\caption{Generation quality of different trigger modalities. CLIPSIM measures image-text semantic alignment ($\uparrow$), and FVD measures video quality ($\downarrow$). Bold denotes the proposed BadAction method.}
\label{tab:quality}
\resizebox{\textwidth}{!}{%
\small
\setlength{\tabcolsep}{5pt}
\begin{tabular}{@{}llccc@{}}
\toprule
Modality Category & Attack Method & Trigger Modality & CLIPSIM (\%) $\uparrow$ & FVD $\downarrow$ \\
\midrule
\multirow{8}{*}{Single-Modality}
& \textcolor{clipgray}{Benign} & \textcolor{clipgray}{--} & \textcolor{clipgray}{85.1} & \textcolor{clipgray}{345.7} \\
& BadNets \citep{gu2019badnets} & Image & 82.5 & 777.8 \\
& Blended \citep{chen2017blended} & Image & 80.3 & 834.2 \\
& Sig \citep{barni2019sig} & Image & 82.4 & 2125.2 \\
& ReFool \citep{liu2020refool} & Image & 77.9 & 1036.0 \\
& WaNet \citep{nguyen2021wanet} & Image & 82.2 & 803.8 \\
& BadNets-T \citep{gu2019badnets} & Text & 81.8 & 758.1 \\
\rowcolor{lightblue}
& \textbf{BadAction (Ours)} & Action & 82.3 & 759.1 \\
\midrule
\multirow{3}{*}{Dual-Modality}
& Image-Text Attack & Image + Text & 82.5 & 2117.8 \\
& Image-Action Attack & Image + Action & 81.6 & 1968.5 \\
& Text-Action Attack & Text + Action & 82.2 & 772.4 \\
\midrule
Triple-Modality
& Tri-modal Attack & Image + Text + Action & 81.4 & 2050.8 \\
\bottomrule
\end{tabular}%
}
\vspace{-10pt}
\end{table}

\subsection{Ablation Studies}

{
\setlength{\textfloatsep}{5pt plus 1pt minus 2pt}
\begin{figure}[!t]

\makebox[\textwidth][l]{%
  \includegraphics[
    width=\textwidth,
    trim=8pt 0 0 0,
    clip
  ]{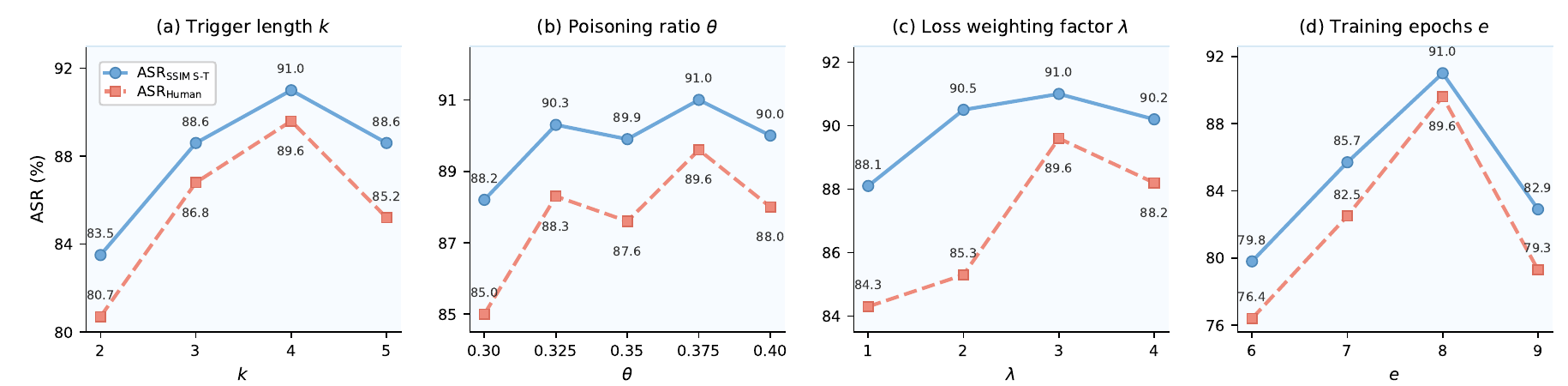}
}

\caption{Ablation studies on the trigger length $k$, the poisoning ratio $\theta$, the loss weighting factor $\lambda$, and the training epochs $e$, using both $\mathrm{ASR}_{\text{SSIM S-T}}$ and $\mathrm{ASR}_{\text{Human}}$ as metrics.}
\label{fig:ablation}

\end{figure}
}

We conduct ablation studies on the trigger length $k$, the poisoning ratio $\theta$, the loss weighting factor $\lambda$, and the training epochs $e$, using both $\mathrm{ASR}_{\text{SSIM S-T}}$ and $\mathrm{ASR}_{\text{Human}}$ as metrics.

\textbf{Effect of Trigger Length.}
To study how the trigger length affects attack effectiveness, we vary
the number of actions $k$ from 2 to 5. Figure~\ref{fig:ablation}(a)
shows that both metrics improve as $k$ increases from 2 to 4 and 
attain their highest values at $k = 4$. Increasing $k$ to 5 reduces both
metrics. This trend is not monotonic, which shows that a longer
trigger does not necessarily yield stronger attack effectiveness. Both
metrics select the same optimum, so we use $k = 4$ in the remaining
experiments.

\textbf{Effect of Poisoning Ratio.}
To evaluate sensitivity to the poisoning ratio, we vary $\theta$ from
0.30 to 0.40. Figure~\ref{fig:ablation}(b) shows that both metrics remain
high across the tested range and vary only modestly. The best performance
occurs at $\theta = 0.375$. The small variation shows that the attack
is not highly sensitive to this hyperparameter within the evaluated
range. We adopt $\theta = 0.375$ in the main experiments.

\textbf{Effect of Loss Weighting Factor.}
We vary $\lambda$ from
1.0 to 4.0. Figure~\ref{fig:ablation}(c) shows that both metrics improve
as $\lambda$ increases from 1.0 to 3.0 and reach their highest values at
$\lambda = 3.0$. A further increase to $\lambda = 4.0$ lowers both
metrics. We therefore adopt $\lambda = 3.0$ in the main
experiments.

\textbf{Effect of Training Epochs.}
We vary the training epoch $e$
from 6 to 9. Figure~\ref{fig:ablation}(d) shows that both metrics improve
as training proceeds to $e = 8$ and then decline at $e = 9$. The two
metrics attain their highest values at $e = 8$. This result shows that
longer training does not necessarily improve backdoor effectiveness. We
 adopt the checkpoint at $e = 8$ in the main experiments.
\vspace{-6pt}
\subsection{Resistance to Existing Defenses}
\vspace{-6pt}

\begin{table*}[t]
\centering

\begin{minipage}[t]{0.48\textwidth}
\centering
\caption{Detection performance of T2IShield and UFID (\%).}
\label{tab:defense}
\resizebox{\linewidth}{!}{%
\begin{tabular}{@{}lccc@{\hspace{0.5em}}ccc@{}}
\toprule
& \multicolumn{3}{c}{\textbf{T2IShield}}
& \multicolumn{3}{c}{\textbf{UFID}} \\
\cmidrule(lr){2-4}
\cmidrule(lr){5-7}
Attack Method
& Pre. & Rec. & F1
& Pre. & Rec. & F1 \\
\midrule
BadNets
& 93.8 & 90.0 & 91.8
& 99.1 & 53.8 & 69.7 \\
Blended
& 94.3 & 100.0 & 97.1
& 98.9 & 45.1 & 62.0 \\
Sig
& 99.4 & 100.0 & 99.7
& 98.5 & 33.9 & 50.4 \\
ReFool
& 99.4 & 100.0 & 99.7
& 98.8 & 42.1 & 59.0 \\
WaNet
& 99.5 & 99.2 & 99.3
& 99.0 & 51.1 & 67.4 \\
BadNets-T
& 99.4 & 85.9 & 92.2
& 99.0 & 48.8 & 65.4 \\
\rowcolor{lightblue}
\textbf{BadAction (Ours)}
& 98.0 & 50.0 & 66.2
& 78.3 & 18.0 & 29.3 \\
\bottomrule
\end{tabular}%
}
\end{minipage}%
\hfill
\begin{minipage}[t]{0.48\textwidth}
\caption{Detection performance of T2IShield and UFID (\%) under different trigger modalities.}
\label{tab:defense_modalities}
\resizebox{\linewidth}{!}{%
\begin{tabular}{@{}lccc@{\hspace{0.7em}}ccc@{}}
\toprule
& \multicolumn{3}{c}{\textbf{T2IShield}}
& \multicolumn{3}{c}{\textbf{UFID}} \\
\cmidrule(lr){2-4}
\cmidrule(lr){5-7}
Trigger Modality
& Pre. & Rec. & F1
& Pre. & Rec. & F1 \\
\midrule
\rowcolor{lightblue}
\textbf{BadAction (Ours)}
& 98.0 & 50.0 & 66.2
& 78.3 & 18.0 & 29.3 \\
Action-Text Attack
& 73.7 & 14.0 & 23.5
& 42.3 & 11.0 & 17.5 \\
Action-Image Attack
& 88.4 & 38.0 & 53.1
& 40.0 & 10.0 & 16.0 \\
Image-Text Attack
& 88.6 & 39.0 & 54.2
& 44.4 & 12.0 & 18.9 \\
Tri-modal Attack
& 88.4 & 38.0 & 53.1
& 30.0 & 15.0 & 20.0 \\
\bottomrule
\end{tabular}%
}
\end{minipage}

\vspace{-8pt}
\end{table*}

We evaluate BadAction against two backdoor detection methods, T2IShield
\citep{wang2024t2ishield} and UFID \citep{guan2025ufid}. Both methods are
originally designed for diffusion models. Detection is performed over 100 backdoored and 100 benign samples at a fixed 5\%
false-positive rate. Table~\ref{tab:defense} shows that both methods achieve
high precision but low recall. T2IShield obtains a precision of 98.0\%, a recall of 50.0\% and an F1 score of 66.2\%. Despite its high
precision, its recall indicates limited separation between backdoored
and benign samples. UFID obtains a
precision of 78.3\%, a recall of 18.0\% and an F1 score of 29.3\% which indicates a weak detection capability for action-based backdoors.\ These results show that existing detectors miss
many action-based backdoors under practical operating points.\ Table~\ref{tab:defense_modalities} shows that multimodal
triggers are less detectable than the action-only trigger by both
detectors, indicating that they provide a more stealthy backdoor setting.
This result highlights the need for greater attention to security in
interactive video generation.
\vspace{-8pt}

\section{Conclusion}
\label{sec:conclusion}
\vspace{-6pt}
We presented the first systematic study of backdoor attacks on
interactive video generation models. Based on this study, we proposed
BadAction, which embeds predefined motion patterns into poisoned action
sequences and pairs them with a static target video. Once triggered, BadAction produces frozen outputs that ignore subsequent user actions while preserving benign behavior. It also supports multimodal triggers by jointly poisoning action, text, and image
channels. Extensive experiments show that BadAction achieves high attack success rates, preserves generation utility, and remains effective against existing backdoor detection methods. We hope this work draws attention to the action control channel as an attack surface and inspires future research on both stronger attacks and dedicated defenses for interactive video generation.

\subsection*{Ethics statement}

This paper studies backdoor attacks on interactive video generation (IVG) models, which may be deployed in safety-critical scenarios such as autonomous driving simulation. Our goal is to expose the vulnerability of the action control channel and to motivate dedicated defenses rather than to facilitate misuse.

\bibliography{iclr2027_conference}
\bibliographystyle{iclr2027_conference}

\end{document}

%% file: math_commands.tex
\usepackage{amsmath,amsfonts,bm}

\def\eqref#1{equation~\ref{#1}}

\def\1{\bm{1}}

\DeclareMathAlphabet{\mathsfit}{\encodingdefault}{\sfdefault}{m}{sl}
\SetMathAlphabet{\mathsfit}{bold}{\encodingdefault}{\sfdefault}{bx}{n}

